\documentclass[conference]{IEEEtran}
\IEEEoverridecommandlockouts
\usepackage{cite}
\usepackage{amsmath,amssymb,amsfonts}
\usepackage{algorithmic}
\usepackage{booktabs}
\usepackage{multicol}
\usepackage{multirow}
\usepackage{graphicx}
\usepackage{textcomp}
\usepackage{xcolor}
\def\BibTeX{{\rm B\kern-.05em{\sc i\kern-.025em b}\kern-.08em
    T\kern-.1667em\lower.7ex\hbox{E}\kern-.125emX}}

\usepackage{scalerel} 
\usepackage{tikz} 
\usetikzlibrary{svg.path} 
\definecolor{orcidlogocol}{HTML}{A6CE39}
\tikzset{
  orcidlogo/.pic={
    \fill[orcidlogocol] svg{M256,128c0,70.7-57.3,128-128,128C57.3,256,0,198.7,0,128C0,57.3,57.3,0,128,0C198.7,0,256,57.3,256,128z};
    \fill[white] svg{M86.3,186.2H70.9V79.1h15.4v48.4V186.2z}
                 svg{M108.9,79.1h41.6c39.6,0,57,28.3,57,53.6c0,27.5-21.5,53.6-56.8,53.6h-41.8V79.1z M124.3,172.4h24.5c34.9,0,42.9-26.5,42.9-39.7c0-21.5-13.7-39.7-43.7-39.7h-23.7V172.4z}
                 svg{M88.7,56.8c0,5.5-4.5,10.1-10.1,10.1c-5.6,0-10.1-4.6-10.1-10.1c0-5.6,4.5-10.1,10.1-10.1C84.2,46.7,88.7,51.3,88.7,56.8z};
  }
}

\newcommand\orcidicon[1]{\href{https://orcid.org/#1}{\mbox{\scalerel*{
\begin{tikzpicture}[yscale=-1,transform shape]
\pic{orcidlogo};
\end{tikzpicture}
}{|}}}}

\usepackage[hidelinks]{hyperref}

\usepackage[nolist,nohyperlinks]{acronym}
\begin{acronym}
\acro{ASMK}{aggregated selective match kernel}
\acro{APE}{absolute positioning error}
\acro{ATE}{absolute trajectory error}

\acro{BA}{bundle adjustment}
\acro{BoW}{Bag-of-Words}
\acro{BRIEF}{binary robust independent elementary features}

\acro{CNN}{convolutional neural network}
\acrodefplural{CNN}[CNNs]{Convolutional neural networks}

\acro{DBoW2}{Bags  of  Binary  Words  for  FAST  Recognition  in  Image  Sequence}
\acro{DOF}{degrees of freedom}

\acro{EKF}{Extended Kalman filter}

\acro{FIM}{Fisher information matrix}

\acrodef{GRU}[GRU]{gated recurrent unit}
\acrodefplural{GRU}{Gated recurrent units}
\acro{GNN}{graph neural network}
\acro{G-CNNs}{Group equivariant Convolutional Neural Networks}

\acro{IMU}{inertial measurement unit}

\acro{KLT}{Kanade-Lucas-Tomasi}

\acro{LIFT}{learned invariant feature transform}
\acro{LSTM}{long short-term memory}
\acrodefplural{LSTM}{long short-term memory networks}

\acro{MSE}{mean squared error}
\acro{NCC}{normalized cross correlation}

\acro{ORB}{Oriented FAST and Rotated BRIEF}

\acro{RANSAC}{random sample consensus}
\acro{RCNN}{recurrent convolutional neural network}
\acro{RNN}{recurrent neural network}
\acrodefplural{RNN}[RNNs]{Recurrent neural networks}
\acro{RPE}{relative position error}

\acro{PCA}{principal component analysis}
\acro{PTAM}{parallel tracking and mapping}

\acro{RMSE}{root-mean-squared error}

\acro{SAD}{sum of absolute differences}
\acro{SfM}{structure from motion}
\acro{SLAM}{simultaneous localization and mapping}
\acro{SSD}{sum of squared differences}
\acro{SVD}{singular value decomposition}

\acro{UUV}{unmanned underwater vehicle}

\acro{VO}{visual odometry}

\end{acronym}
\begin{document}

\title{Loss it right: Euclidean and Riemannian Metrics in Learning-based Visual Odometry
}

\author{
Olaya~Álvarez-Tuñón \orcidicon{0000-0003-3581-9481},
Yury Brodskiy \orcidicon{0009-0002-0445-8126}, 
and Erdal Kayacan \orcidicon{0000-0002-7143-8777}
    \thanks{
        O. Álvarez-Tuñón is with Artificial Intelligence in Robotics Laboratory (AiRLab), the Department of Electrical and Computer Engineering, Aarhus University, 8000 Aarhus C, Denmark
                {\tt\small \{olaya at ece.au.dk\}}.
        Y. Brodskiy is with EIVA a/s, 8660 Skanderborg, Denmark {\tt\small \{ybr at eiva.com\}}. 
        E. Kayacan is with Automatic Control Group (RAT), Paderborn University, 33098 Paderborn, Germany {\tt\small \{erdal.kayacan at uni-paderborn.de\}}.
        
    }%
}

\maketitle

\begin{abstract}
This paper overviews different pose representations and metric functions in visual odometry (VO) networks. The performance of VO networks heavily relies on how their architecture encodes the information. The choice of pose representation and loss function significantly impacts network convergence and generalization. We investigate these factors in the VO network DeepVO by implementing loss functions based on Euler, quaternion, and chordal distance and analyzing their influence on performance. The results of this study provide insights into how loss functions affect the designing of efficient and accurate VO networks for camera motion estimation. The experiments illustrate that a distance that complies with the mathematical requirements of a metric, such as the chordal distance, provides better generalization and faster convergence.
The code for the experiments can be found at \url{https://github.com/remaro-network/Loss_VO_right}
\end{abstract}


\section{Introduction}

Several deep learning architectures have arisen in the last decade to support learning from spatio-temporal data, which have allowed the implementation of architectures for end-to-end \ac{VO} methods \cite{surveyDLspatialdata}. PoseNet \cite{kendall2015posenet} introduced end-to-end visual localization with a \ac{CNN} pretrained for classification. Later architectures introduced optical flow networks followed by \ac{CNN} pose regressors \cite{wang2017deepvo, wang2020tartanvo, costante2018lsvo} or 3D CNNs \cite{dl:vo:3dcvo}.
DeepVO \cite{wang2017deepvo} models the temporal relationships between sequences of images with two \ac{LSTM} units. AtLoc \cite{dl:vo:wang2020atloc} incorporates contextual information as a self-attention mechanism operating in both the spatial and temporal domains. TartanVO \cite{wang2020tartanvo} aims for better generalization by inputting the intrinsic camera parameters into the network.
Within the literature, there are discussions on how the inductive bias of a network can impact the information encoded by the model \cite{bronstein2021geometricDL,dl:vo:19cnnlimitations}. For CNNs, their inductive bias is shift equivariance \cite{dl:vo:cotogni2022offset}. Recent research has shown that incorporating translation and rotation-equivariant features can reduce the required data and improve generalization by introducing motion representation into the feature space \cite{dl:vo:equivfeaturesforposeregression, dl:vo:brynte2022rigidity}.

In addition to the architecture, the choice of loss function significantly impacts the information encoded by the network. In \ac{VO}, the main challenge to address when implementing loss functions is the rotation representation \cite{carlone2015duality}.
The supervised pose regression networks usually rely on the Euclidean loss for rotation and translation \cite{kendall2015posenet, wang2017deepvo,dl:vo:wang2020atloc}. These methods implement different variations of a weighted sum of the Euclidean loss for the rotation and translation components.
PoseNet and Atloc represent rotations with quaternions. DeepVO outputs Euler angles to avoid the unit constraint restriction on the optimization. Quaternions and Euler angles define a discontinuous representation for rotations, a source of error for neural networks. On the contrary, the Lie algebra representation leverages an unconstrained optimization problem, and therefore has recently gained popularity on VO methods \cite{wang2020tartanvo, fang2022exploiting, de2020simultaneously}. 

This paper overviews different pose representations and the metric functions that can be applied to them. 
With the state-of-the-art \ac{VO} network DeepVO as the backbone, we study the influence of pose representation and loss function on the network's convergence and generalization ability.



\section{Geometry for visual odometry}

\begin{figure}[b]

    \centering
    \includegraphics[width=0.8\columnwidth]{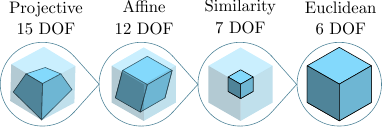}
    \caption{Hierarchy of geometries. Euclidean preserves areas, angles and lengths, the similarity preserves ratios of lengths. The affine preserves volumetric ratios and parallelism. The projective preserves intersections and tangents.}
    \label{fig:intro:geometries}
\end{figure}

The geometry representation that best models the perspective projection from camera imaging is projective geometry. 
Projective transforms only preserve type (points or lines), incidence (whether a point lies in a line) and cross-ratio. Euclidean geometry is a subset of projective geometry that, in addition to that, also preserve lengths, angles and parallelism. There are two other hierarchies between them: affine and similarity geometry, as shown in Fig. \ref{fig:intro:geometries}.



Projective reconstruction leads to distorted models; thus visual localization algorithms aim to estimate the Euclidean structure of the scene \cite{hartleyzisserman2003multiple}. Moreover, \ac{VO} requires quantifying differences in translation and orientation between camera frames.
The distance between two points, represented as vectors in Euclidean space, is measured by the Euclidean distance,
which measures the straight-line length between them. However, \ac{VO} requires quantifying differences in translation and orientation between camera frames.
Measuring differences in orientation between two camera poses in Euclidean space requires more complete representations of relative positions and orientations of objects in space. The next section introduces the different pose representations used in Euclidean geometry and their role in the pose optimization problem.

\subsection{Pose parameterization and optimization}
Pose representation involves translation and rotation with respect to a reference frame, which can be parameterized using Euler angles, quaternions, or rotation matrices.
Euler angles are often used for their intuitiveness. The pose is represented by the 3D translations
plus the special case of Euler angles yaw-pitch-roll, thus defining the 6D vector:
    \begin{equation}
        P_6 = \left[t\text{ }\varphi\right]^T =\left[x\text{ }y\text{ }z\text{ }\phi\text{ }\chi \text{ }\psi\right]^T
    \end{equation}
However, they can lead to numerical instabilities caused by gimbal lock. Additionally, Euler angles are non-linear, which makes the optimization more difficult.
Quaternion is a better representation when compared to Euler angles which avoids the gimbal lock problem and is more numerically stable. The quaternion parameterization has three degrees of freedom and is represented by four elements, with $q_w$ the unit length constraint:
    \begin{equation}
        P_7 = \left[t\text{ }q \right]^T = \left[x\text{ }y \text{ }z \text{ }q_w\text{ } q_x \text{ }q_y \text{ }q_z \right]^T
    \end{equation}
Quaternions are compact and efficient to manipulate mathematically. However, they double-cover the space of rotations, with $q$ and $-q$ yielding the same rotation. 

The rotation matrices belong to the special orthogonal group $SO(3) = \{ R \in \mathbb{R}^{3\times 3}: R^TR = I_3, det(R) = 1 \}$, and their combination with the position information form the special Euclidean group $SE(3) = \{ T = (R, t) \in \mathbb{R}^{4\times 4} \mid R \in SO(3), t \in \mathbb{R}^3 \}$.
:
    \begin{equation}
        T_{4\times 4} =
            \left[ 
                \begin{array}{c|c}
                R & t \\ \hline
                \mathbf{0} & 1
            \end{array} \right]
    \end{equation}
with $R \in SO(3)$ is the rotation matrix  and t the translation vector $t = [x\text{ } y\text{ } z]^T$. 
Rotation and transformation matrices can be considered elements of a Lie group forming a smooth manifold, forming a continuous set of elements that can be smoothly parameterized.


\section{Distances in the Euclidean space} 
Learning-based VO requires finding an appropriate metric that can accurately and robustly quantify the quality of the estimated camera motion. 
In mathematics, a metric can be defined on any set $d: X \times X \rightarrow \mathbb{R} $. A metric or distance $d(A,B)$ between two elements $A$ and $B$ of a certain set satisfies the following properties:
\begin{align}
    \label{eqs:properties}
    \textrm{non-negativity: }& d(A,B) \ge 0 \\
    \textrm{identity: }& d(A,B) = 0 \Leftrightarrow A = B \\
    \textrm{symmetry: }& d(A,B) = d(B,A) \\
    \textrm{triangle inequality: }& d(A,C) \leq d(A,B)+d(B,C)
\end{align}
 In differential geometry, a metric is specifically defined on a geometric space (e.g. a Euclidean space or a Riemannian manifold), and thus it reflects the underlying geometric structure of the space.
 
Translations have a straightforward geometric interpretation as displacements in space. 
On the other hand, rotations do not have a direct geometric interpretation and require other mechanisms to find a metric.
A Riemannian manifold defines a smooth space equipped with a Riemannian metric on the tangent space, which varies smoothly. 
The Euclidean space is a special case of a Riemannian manifold, with a constant curvature of zero. 
Therefore, the following section discusses prevalent metrics and pseudo-metrics used in the literature for the proposed parameterizations for orientations.

\subsection{Distances in the vector space of orientations}
Some well-known distances used for 3D rotations in the VO and SLAM literature are the Euclidean distance, the geodesic distance and the chordal distance, applied to the different orientation parameterizations as follows \cite{carlone2015initialization}:
\subsubsection{Distance on Euler angles} two three-dimensional vectors of Euler angles $\varphi_A$ and $\varphi_B$ yield the Euclidean distance as:  
\begin{equation}
    d_e(\varphi_A,\varphi_B) = ||\varphi_A - \varphi_B||_2^2 
\end{equation}
with $||\cdot||_2$ denoting the $L_2$ norm. 
However, this distance cannot be considered a metric as it does not satisfy the triangle inequality. This induces singularities during the optimization problem, potentially leading to suboptimal solutions or slow convergence.

\subsubsection{Quaternion distance}
Two unit quaternions $q_A$ and $q_B$ yield the Euclidean distance:
\begin{equation}
    d_q(q_A,q_B) = \|q_A-q_B\|_2^2 
\end{equation}
However, due to the quaternion's double-cover, $q_B$ and $-q_B$ represent the same rotation but do not retrieve the same distance, i.e.:
    $d(q_A,q_B) \neq d(q_A,-q_B)$.
This matter can be addressed by redefining the quaternion distance as:
\begin{equation}
\label{eq:quaterniondistance}
    d_q(q_A,q_B) = min_{b \in\{-1;+1\}} \|q_A-bq_B\| 
\end{equation}
which defines a pseudo-metric since it does not satisfy the identity property. Furthermore, it incorporates the problem of adding a binary variable to the equation, hindering the computational analysis. 
\subsubsection{Geodesic distance}
The distance between two rotations $R_A \in SO(3)$ and $R_B \in SO(3)$ can be obtained as the rotation angle $\theta_{AB}$ corresponding to the relative rotation $R_{AB}=R^T_AR_B$:
\begin{equation}
    d_\theta(R_A,R_B) = \|arccos\left(\frac{tr\left(R^T_AR_B\right)-1}{2}\right)\|
\end{equation}
The norm of the exponential coordinates is the rotation angle; thus, the previous metric can be written as:
\begin{equation}
\label{eq:so3:geodesic}
    d_\theta(R_A,R_B) = \|log(R^T_AR_B)^\vee\| = \|log(R^T_BR_A)^\vee\|
\end{equation}
This distance is geodesic, i.e., the length of the minimum path between $R_A$ and $R_B$ on the $SO(3)$ manifold. 
 The geodesic distance defines a Riemannian metric that satisfies the metric properties. Moreover, it defines a smooth metric since both the logarithmic map and the Euclidean norm are smooth. 
However, it brings more computational expense and numerical instability from the logarithm map for small rotations.

\subsubsection{Chordal distance}
\begin{figure}[b]

    \centering
    \includegraphics[width=0.8\columnwidth]{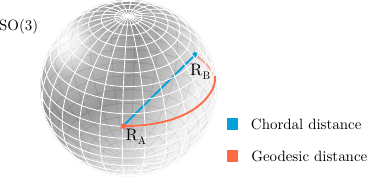}
    \caption{Relationship between the chordal distance and the geodesic distance in the SO(3) sphere for the two rotations $R_A$ and $R_B$.The chordal distance is the straight-line distance between the two points. The geodesic distance is the shortest curve along the sphere's surface connecting the two points on the manifold. The geodesic distance is always greater than or equal to the chordal distance. }
    \label{fig:chordalvsgeo}
\end{figure}

The chordal distance between two rotations $R_A \in SO(3)$ and $R_B \in SO(3)$ is defined as \cite{carlone2015duality}: 
\begin{equation}
\label{eq:so3:chordal}
    d_c(R_A,R_B) = \|R_A-R_B\|_F = \|R_AR_B^T-I\|_F
\end{equation}
with $\|\cdot\|_F$ the Frobenius norm.
In most applications, minimising the square of the chordal distance is preferred. The chordal distance is not a Riemannian metric, but it complies with the four metric requirements while being more numerically stable and simpler than the geodesic distance \cite{so3:chordalvsgeodesic}. The relationship between the chordal and geodesic distance is shown in Fig. \ref{fig:chordalvsgeo}. 


\subsection{Distances in the vector space of linear transformations}
Translations can be measured by satisfying the properties of a metric with the Euclidean distance. However, that property does not hold for transformations.
Distance functions for full rigid-body motions present several implementation challenges: First, no continuous metric functions are invariant under the arbitrary left and right displacements. That is, existing smooth distance functions for SE(3) are invariant under either a left or a right displacement by group elements, but not both \cite{se3:partialbiinvariance}. Second,  since rotations and translations are measured in different units, combining both into one number is not straightforward.

When the transformation is expressed as a vector of translation and orientation with Euler angles or quaternions, the distances are implemented as a weighted sum of the Euclidean distances for the rotation and the translation part. This yields the limitations as mentioned earlier inherent to distances in orientations.
In SE(3), the chordal distance for two transformation matrices $T_A$ and $T_B$ can be generalized from Eq. (\ref{eq:so3:chordal}) as:
\begin{equation}
    d_c(T_A,T_B) = \|T_A-T_B\|_F = \|T_B-T_A\|_F
\end{equation}
It can be proved that \cite{carlone2015duality}:
\begin{equation}
    d_c(T_A,T_B) = \sqrt{d_c(R_A,R_B)^2+\|t_B-R_Bt_A\|^2}
\end{equation}

\section{Experiments}
Considering the distances introduced above, this section proposes a series of experiments to compare them.
\begin{figure*}[ht]
    \centering
    \includegraphics[width=\textwidth]{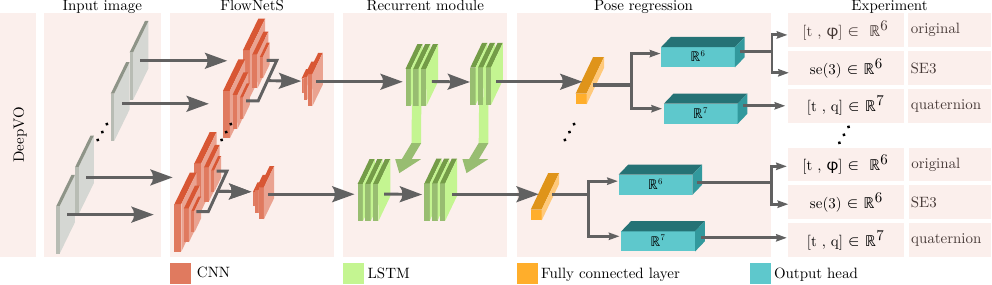}
    \caption{DeepVO's \cite{wang2017deepvo} architecture and output shape according to each experiment. The tensor's output shape in the original setup corresponds to six values for the translation vector and the Euler angles. The SE(3) experiment provides the same output shape, in this case interpreted as the Lie algebra vector se(3). Finally, the translation vector concatenated with the quaternion vector retrieves an output with 7 values.}
    \label{fig:experimentsetup}
\end{figure*}
\subsection{Experiment setup}
The network chosen to investigate the influence of the pose loss functions in the model's performance is DeepVO \cite{wang2017deepvo}. 
DeepVO yields a pre-trained FlowNet \cite{dosovitskiy2015flownet} \ac{CNN} preceded by two \ac{LSTM} units. 
DeepVO is well suited for the proposed experiment since it only requires a loss purely dependent on the distance between the target and the estimated pose.


Figure \ref{fig:experimentsetup} displays the setup for the  three experiments conducted to train DeepVO using different orientation parameterizations: Euclidean angles, quaternions, and the SE(3) pose. All three experiments were carried out under the same training setup.

The loss function originally proposed in DeepVO corresponds to the \ac{MSE} of the Euclidean loss, with the orientation represented as Euler angles:
\begin{equation}
    L_{original} = \frac{1}{N}\sum_{i=1}^{N}\sum_{j=1}^{M}||t_i - \hat{t}_i||_2^2 + k_1||\varphi_i - \hat{\varphi}_i||_2^2
\end{equation}
where for each observation $i$, $\hat{t}_i$ and $\hat{\varphi}_i$ are the estimated values for rotation and orientation with ground truth $t_i$ and $\varphi_i$.  $M$ and $N$ correspond to the sequence length and the number of observations, respectively.

Similarly, the Euclidean loss for the pose represented as translation and quaternion vectors is obtained as:
\begin{equation}
    L_{quat} = \frac{1}{N}\sum_{i=1}^{N}\sum_{j=1}^{M}||t_i - \hat{t}_i||_2^2
    + \underset{b \in\{-1;+1\}}{min} k_2||q_i - b\hat{q}_i||_2^2
\end{equation}
with $q_i$ and $\hat{q}_i$ the ground truth and estimated quaternions. For this experiment, the output head of the network was modified to provide seven outputs corresponding to the three translation elements and the four rotation elements.

Finally, the chordal loss for the SE(3) representation is obtained as:
\begin{equation}    L_{SE(3)} = \frac{1}{N}\sum_{i=1}^{N}\sum_{j=1}^{M}\|t_i-\hat{t}_i\|_2^2+k_3d_c(R_i,\hat{R}_i)^2
\end{equation}
with $R_i$ and $\hat{R}_i$ the ground truth and estimated SO(3) rotations. It is to be noted that the rotation matrix is obtained from the network's output vector through the exponential map.

The three constants $k_1$,$k_2$ and $k_3$ are experimentally determined. They aim to equate the order of magnitude of the orientations to that of the translations. For these experiments, $k_1 = 100$, $k_2 = 14$, and $k_3 = 153$.

The training is carried out with the KITTI \ac{VO} dataset \cite{dataset:kitti} with the original image sizes. We select sequences 00, 02-06, and 08 as training sets and sequences 09 and 07 as validation sets. With a pre-trained FlowNet as an initial checkpoint, the network is trained for up to 200 epochs with a 20\% dropout and using an Adam optimizer with a learning rate of 0.001, as originally proposed by DeepVO \cite{wang2017deepvo}. The selected model corresponds to the epoch that provides the lowest error in the validation set.

\subsection{Results and discussion}
\begin{figure}[ht]
    \centering
    \includegraphics[width=\columnwidth]{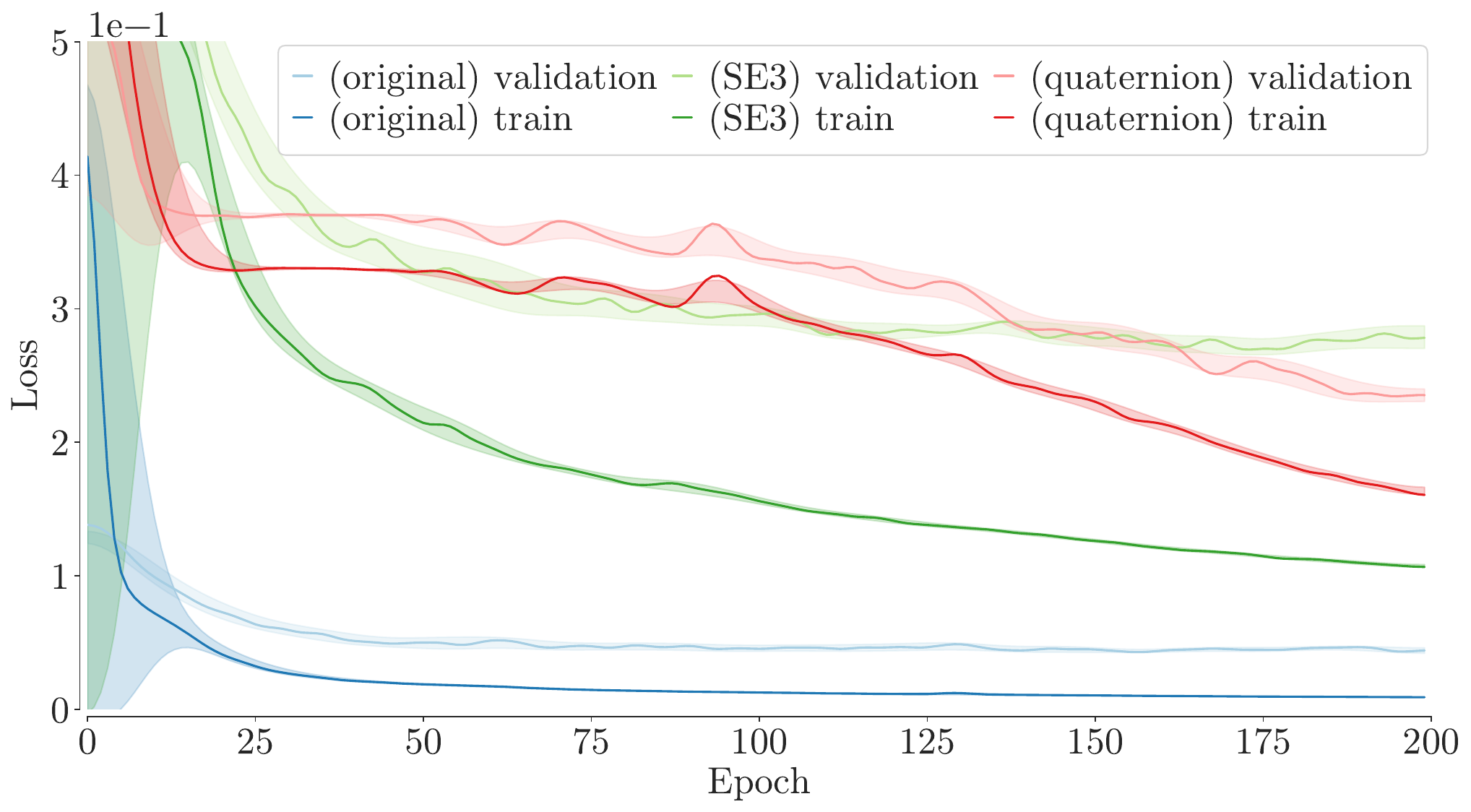}

    \includegraphics[width=.49\columnwidth]{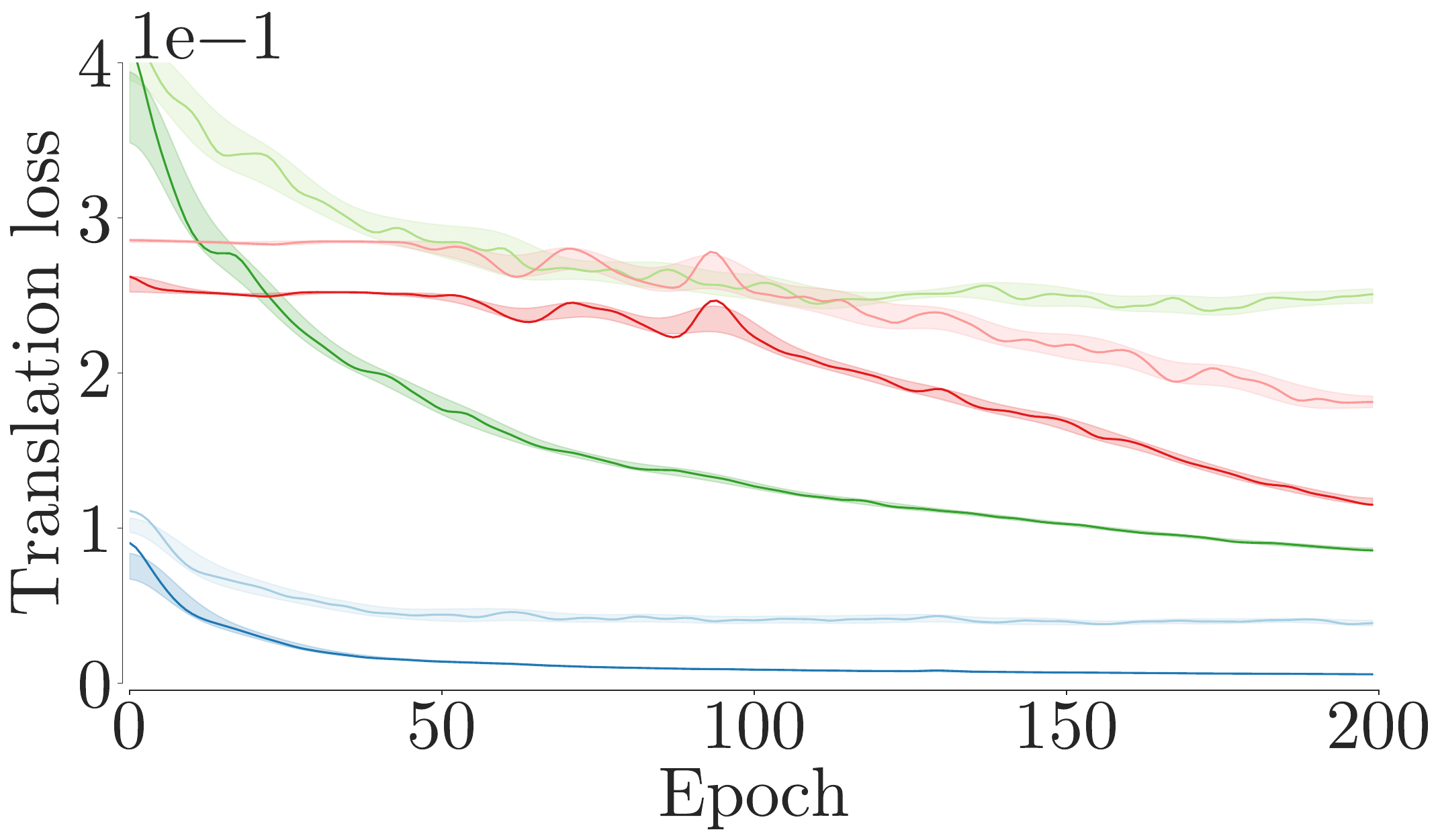}
    \includegraphics[width=.49\columnwidth]{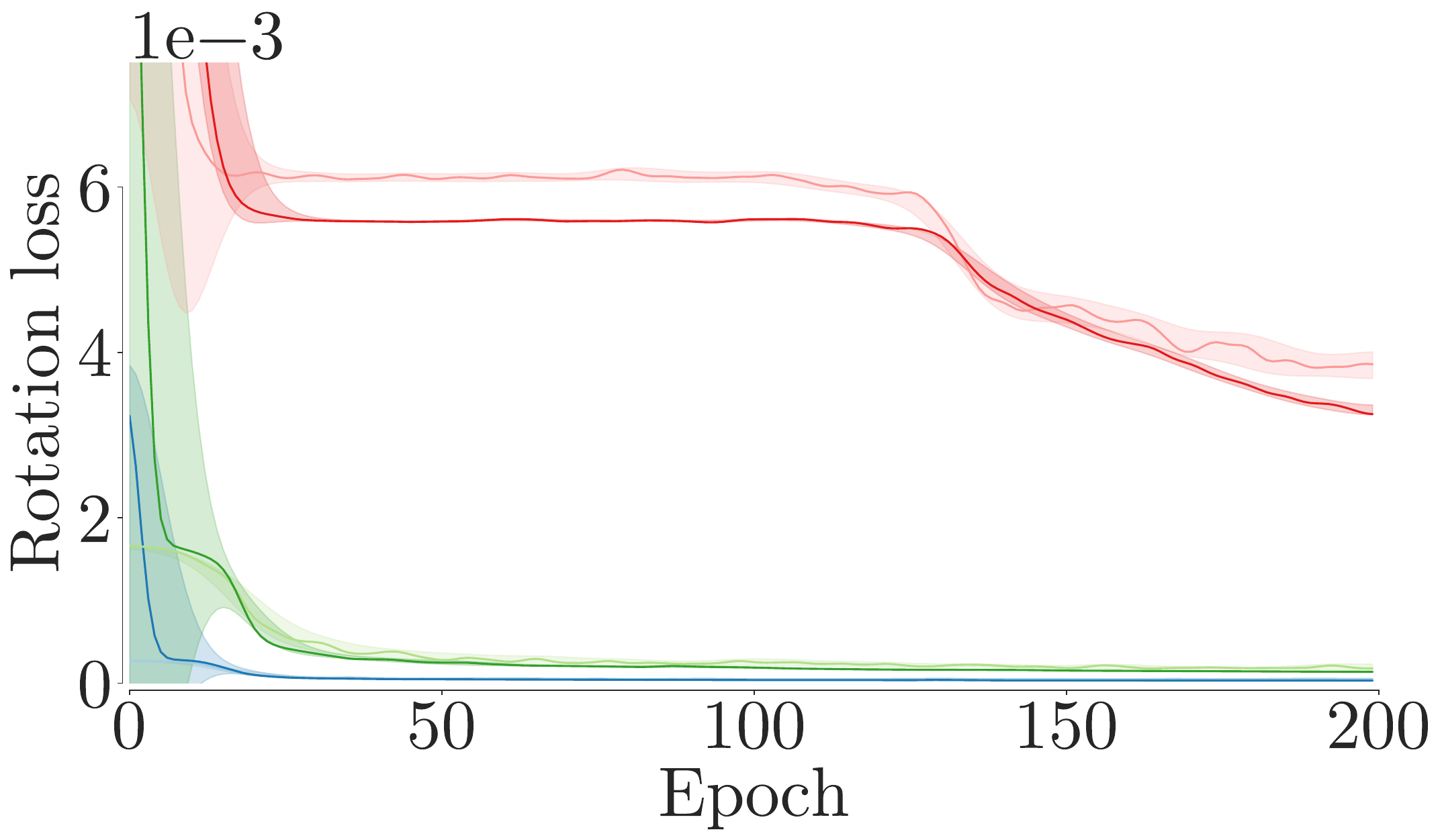}
    
    \caption{Top: pose loss values for the pose loss during train and validation. Bottom: Translation and rotation losses (without weighting). Note that the rotation losses do not have the same geometric interpretation.}
    \label{fig:trainvalresults}
\end{figure}

The train and validation losses for the three experiments are shown in Fig. \ref{fig:trainvalresults}. First, Fig. \ref{fig:trainvalresults} illustrates a lack of data for adequate convergence. Due to time limitations, feeding more data to the network and tuning the regularization parameters are left as future works. Instead, these experiments are taken as a pure comparison of the losses' performance under the same conditions. 
Figure \ref{fig:trainvalresults} shows a better convergence of the loss function under the chordal loss from the SE(3) parameterization, evidenced by a steady decrease over the 200 epochs. This decrease is present in training and validation, which indicates better generalization. As opposed to that, the loss under the Euler angle parameterization presents a rapid but mild decrease of the loss for the first epochs during training, then enter a flat area. The loss under the quaternion parameterization shows a slow convergence where it seems to converge to a local minimum between epochs 25 and 100, then steadily decrease again.  

\begin{figure}[ht]

    \includegraphics[width=.32\columnwidth]{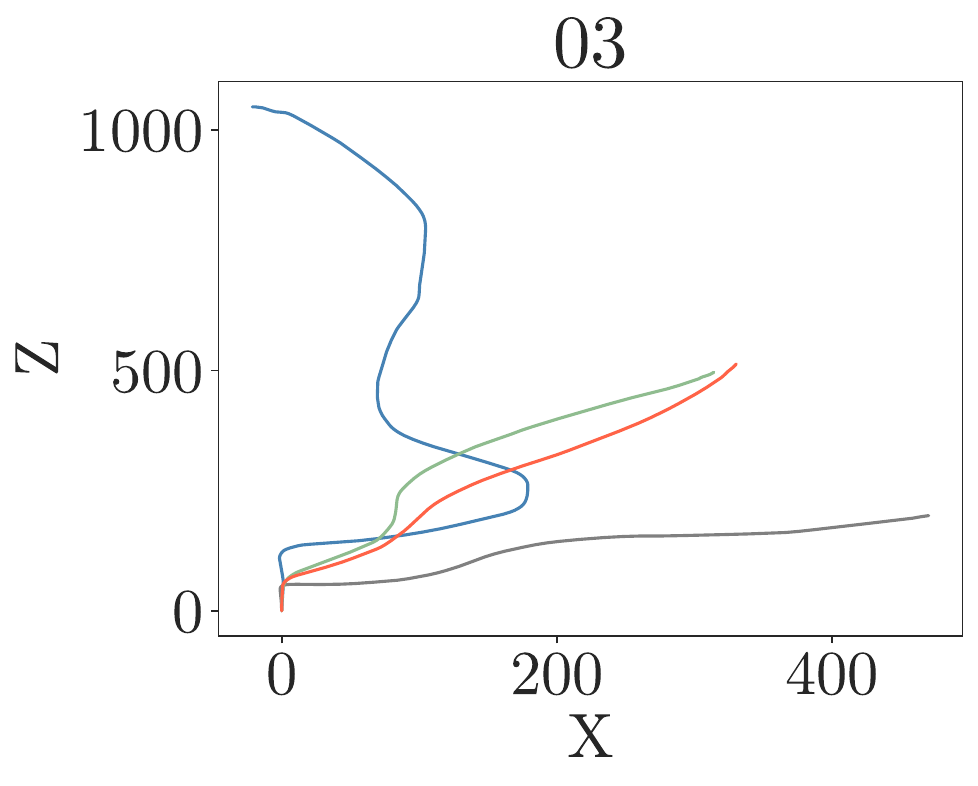} 
    \includegraphics[width=.32\columnwidth]{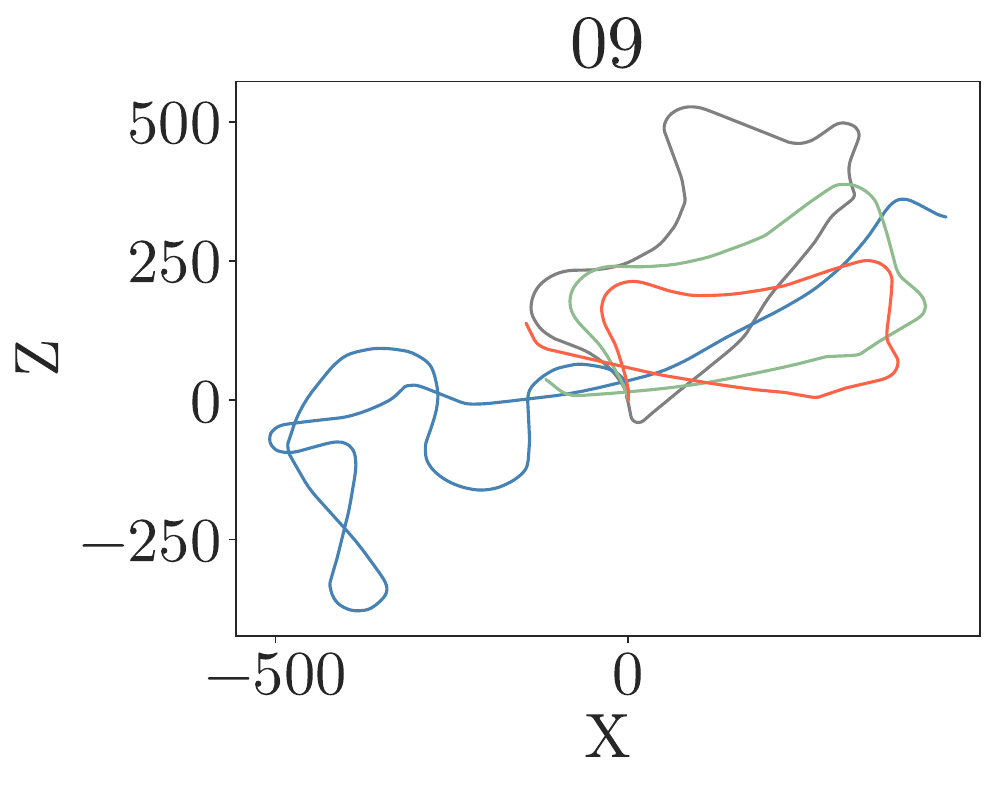} 
    \includegraphics[width=.32\columnwidth]{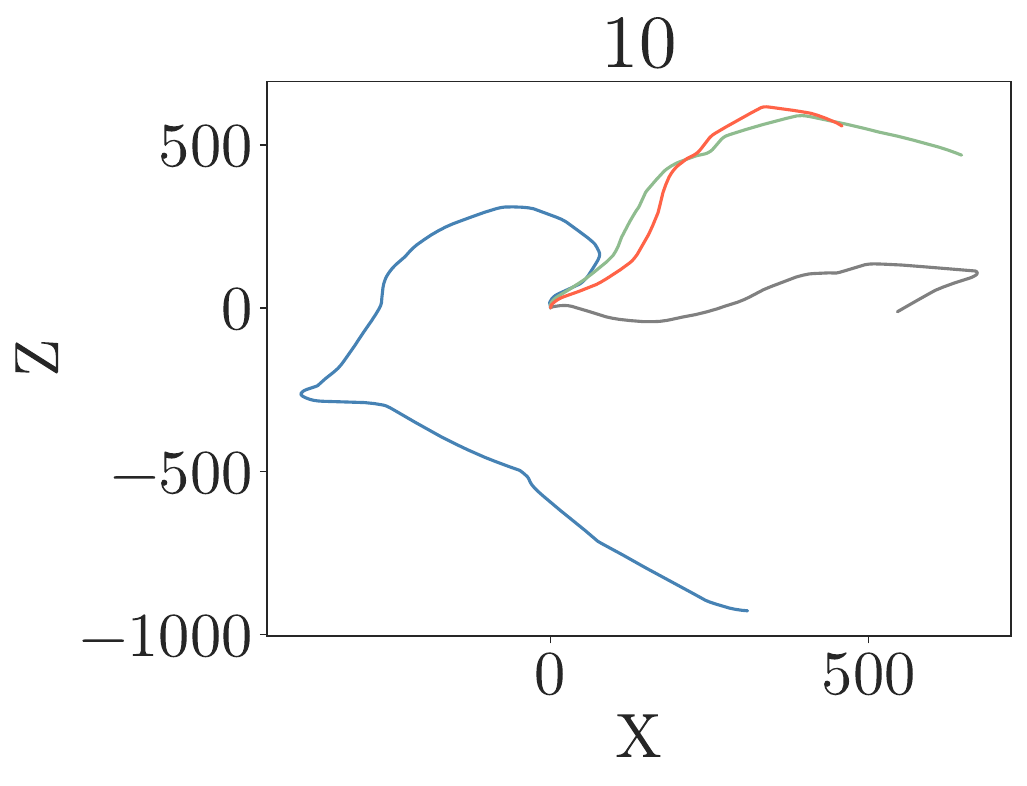}
    \includegraphics[width=\columnwidth]{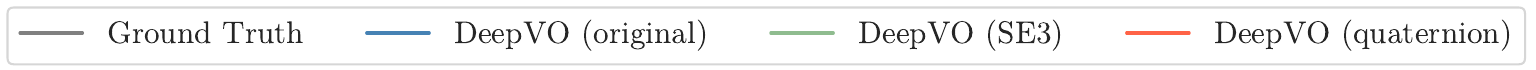}
    
    \caption{
    From left to right, trajectories used for training, validation, and test. The plot shows the ground truth versus estimated pose for DeepVO using the proposed pose representations and loss functions.}
    \label{fig:traintraj}
\end{figure}

The errors from the results inferred by each experiment's model are depicted in Table \ref{table:transerror} and \ref{table:roterror}. The local and global accuracies are quantified with the \ac{APE} and the \ac{RPE} between the ground truth and estimate as implemented in \cite{grupp2017evo}. An example of the estimated trajectories is shown in Fig. \ref{fig:traintraj}.
The results show the superiority of the models trained with quaternions and SE(3) representation versus the original implementation. The Euler angles representation presents big rotation shifts even in the data seen during training. The models using quaternions and SE(3) representation show similar performance, although the SE(3) model presents a slightly better performance in most trajectories. Figure \ref{fig:traintraj} evidences big orientation drifts under steep rotations on the Trajectories 03 and 10. This is due to the low presence of steep rotations in the dataset, which hinders learning such data distributions.
In conclusion,  the SE(3) representation with a chordal loss presents the best performance under the fastest convergence.

\section{Conclusions}
The findings of this study indicate that using chordal loss with SE(3) pose representation is promising for encoding the underlying geometry of \ac{VO}. Compared to the Euclidean loss with quaternion and Euler angles parameterization, this approach has the potential to result in improved convergence and generalization. One reason is that the chordal loss can account for the nonlinearity of the manifold of rotations. Furthermore, the chordal loss satisfies the mathematical metric requirements while being more numerically stable than other Riemannian metrics. Overall, these findings suggest that using chordal loss with SE(3) pose representation is a valuable tool for improving the accuracy and robustness of VO. The implications of this realization are twofold: firstly, the loss function strongly influences the information encoded by the network independently of the architecture. Secondly, in addition to the data availability, the choice of loss function conditions its ability to converge and generalize: a loss that is representative of the geometric space will allow the network to keep learning on the data throughout the epochs.

Future works include adding more data and performing hyperparameter optimization to obtain a better fit for the model and the implementation of more losses in the Riemannian and Euclidean spaces.



\begin{table}[ht!]
\centering
\footnotesize
\caption{Translation errors from each experiment's model. The best results are marked in \textbf{bold}.
}
\label{table:transerror}
\begin{tabular}{c  cc cc cc }
\toprule
\multirow{2}{*}[-1.2em]{}  & \multicolumn{2}{c}{DeepVO(original)} & \multicolumn{2}{c}{DeepVO(SE(3))} & \multicolumn{2}{c}{DeepVO(quaternion)}\\ 
\cmidrule(l{1.5em}r{1.5em}){4-5}
\cmidrule(l{1.5em}r{1.5em}){6-7}
\cmidrule(l{1.5em}r{1.5em}){2-3}
& APE[m] & RPE[m]  &  APE[m] & RPE[m] &  APE[m] & RPE[m] \\
\toprule

00       & 1345.5 & .884   & \textbf{1110.1}  & .154 & 1165.4 & \textbf{.147} \\ 
01       & 1389.8 & \textbf{.789}   & \textbf{1233.5}  & 1.36 & 1353.9 & 1.45 \\ 
02       & 1467.9 & 1.09   & \textbf{1132.6}  & \textbf{.134} & 1165.5 & .154 \\
03       & 585.1 & .922   & \textbf{205.1 } & .189 & 212.48 & \textbf{.165} \\
04       & 206.7 & 1.17   & \textbf{16.1}  & \textbf{.108} & 19.1 & .141 \\
05       & 720.1 & .886   & \textbf{313.2}  & \textbf{.137} & 504.7 & .151 \\
06       & 820.8 & 1.09   & \textbf{442.2}  & \textbf{.118} & 519.4 & .184 \\
07       & \textbf{322.9} & .952   & 371.1  & \textbf{.285} & 478.9 & .289 \\
08       & 1021.0 & .902  & \textbf{522.8}  & .\textbf{147} & 879.2 & .173 \\
09       & 614.3 & .882   & \textbf{180.9}  & \textbf{.243} & 342.2 & .257 \\
10       & 703.5 & 1.08   & \textbf{421.9}  & .321 & 465.9 & \textbf{.309} \\

\bottomrule
\end{tabular}
\end{table}

\begin{table}[ht!]
\centering
\footnotesize
\caption{Orientation errors from each experiment's model. The best results are marked in\textbf{ bold}.
}
\label{table:roterror}
\begin{tabular}{c  cc cc cc }
\toprule
\multirow{2}{*}[-1.2em]{}  & \multicolumn{2}{c}{DeepVO(original)} & \multicolumn{2}{c}{DeepVO(SE(3))} & \multicolumn{2}{c}{DeepVO(quaternion)}\\ 
\cmidrule(l{1.5em}r{1.5em}){4-5}
\cmidrule(l{1.5em}r{1.5em}){6-7}
\cmidrule(l{1.5em}r{1.5em}){2-3}
& ATE[rad] & RPE[rad]  &  ATE[rad] & RPE[tad] &  ATE[tad] & RPE[rad] \\
\toprule

00       & \textbf{1.91} & .025   & 2.35  & \textbf{.020} & 2.49 & .024 \\
01       & \textbf{1.68} & .018   & 1.78  & .015 & 2.00 & \textbf{.012} \\
02       & \textbf{1.66} & .021   & 2.15  & \textbf{.015} & 2.45 & .017 \\
03       & 1.89 & .020   & \textbf{.910}  & .012 & 1.16 & \textbf{.010} \\
04       & .734 & .0085   & \textbf{.164}  & \textbf{.0047} & .411 & .0049 \\
05       & 1.91 & .022  & \textbf{1.13}  & .013 & 1.64 & \textbf{.012} \\
06       & 2.20 & \textbf{.017}   & \textbf{2.11}  & \textbf{.017} & 2.03 & .021 \\
07       & \textbf{1.93} & .029   & 2.03  & \textbf{.018} & 2.03 & .024 \\
08       & 1.76 & .022  & \textbf{1.29}  & \textbf{.014} & 2.10 & .016 \\
09       & 2.52 & .023   & \textbf{1.02}  & \textbf{.014} & 1.59 & \textbf{.014} \\
10       & 2.17 & .027   & \textbf{1.63}  & .017 & 1.78 & \textbf{.015} \\

\bottomrule
\end{tabular}
\end{table}

\section*{Acknowledgment}
This paper is written under the project REMARO which has received funding from the European Union's EU Framework Programme for Research and Innovation Horizon 2020 under Grant Agreement No 956200. 

\bibliographystyle{IEEEtran}
\bibliography{References}

\end{document}